\documentclass[final,5p,times,twocolumn,authoryear]{elsarticle}

\usepackage{amssymb}
\usepackage{amsmath}
\usepackage{array}
\usepackage{algorithmic}
\usepackage{graphicx}
\usepackage{booktabs}
\usepackage{multirow}
\usepackage{pifont}
\usepackage{makecell}
\newcommand{\cmark}{\ding{51}}%
\newcommand{\xmark}{\ding{53}}%

\usepackage[colorlinks,citecolor=blue,urlcolor=blue]{hyperref}

\journal{}

\begin{document}

\begin{frontmatter}



\title{Learning to Efficiently Adapt Foundation Models for Self-Supervised Endoscopic 3D Scene Reconstruction from Any Cameras} 


\author[CUHK]{Beilei Cui\fnref{eqcon}}
\ead{beileicui@link.cuhk.edu.hk}

\author[CUHK,TUM]{Long Bai\fnref{eqcon}}
\ead{b.long@link.cuhk.edu.hk}

\author[UCL]{Mobarakol Islam\fnref{eqcon}}
\ead{mobarakol.islam@ucl.ac.uk}

\author[CUHK]{An Wang}
\ead{wa09@link.cuhk.edu.hk}

\author[CUHK]{Zhiqi Ma}
\ead{zhiqima@link.cuhk.edu.cn}

\author[CUHK]{Yiming Huang}
\ead{yhuangdl@link.cuhk.edu.hk}

\author[TUM]{Feng Li}
\ead{feng.li@tum.de}

\author[Yale]{Zhen Chen}
\ead{zchen.francis@gmail.com}

\author[TUM]{Zhongliang Jiang}
\ead{zl.jiang@tum.de}

\author[TUM]{Nassir Navab}
\ead{nassir.navab@tum.de}

\author[CUHK]{Hongliang Ren\corref{cor1}}
\ead{hlren@ieee.org}

\fntext[eqcon]{Co-first authors}
\cortext[cor1]{Corresponding author}
\affiliation[CUHK]{organization={Department of Electronic Engineering, The Chinese University of Hong Kong},
            city={Hong Kong},
            country={Hong Kong}}          

\affiliation[TUM]{organization={Chair for Computer Aided Medical Procedures, Technical University of Munich},
            city={Munich},
            country={Germany}}
            
\affiliation[UCL]{organization={UCL Hawkes Institute, University College London},
            city={London},
            country={United Kingdom}}

\affiliation[Yale]{organization={Yale University},
            city={New Haven},
            country={USA}}


\begin{abstract}
Accurate 3D scene reconstruction is essential for numerous medical tasks. Given the challenges in obtaining ground truth data, there has been an increasing focus on self-supervised learning (SSL) for endoscopic depth estimation as a basis for scene reconstruction. While foundation models have shown remarkable progress in visual tasks, their direct application to the medical domain often leads to suboptimal results. However, the visual features from these models can still enhance endoscopic tasks, emphasizing the need for efficient adaptation strategies, which still lack exploration currently. In this paper, we introduce Endo3DAC, a unified framework for endoscopic scene reconstruction that efficiently adapts foundation models. We design an integrated network capable of simultaneously estimating depth maps, relative poses, and camera intrinsic parameters. By freezing the backbone foundation model and training only the specially designed Gated Dynamic Vector-Based Low-Rank Adaptation (GDV-LoRA) with separate decoder heads, Endo3DAC achieves superior depth and pose estimation while maintaining training efficiency. Additionally, we propose a 3D scene reconstruction pipeline that optimizes depth maps' scales, shifts, and a few parameters based on our integrated network. Extensive experiments across four endoscopic datasets demonstrate that Endo3DAC significantly outperforms other state-of-the-art methods while requiring fewer trainable parameters. To our knowledge, we are the first to utilize a single network that only requires surgical videos to perform both SSL depth estimation and scene reconstruction tasks. The code will be released upon acceptance.
\end{abstract}


\end{frontmatter}



\section{Introduction}
\label{sec:introduction}

Surgical 3D scene reconstruction holds immense value in minimally invasive surgery for improving the effectiveness of VR/AR-assisted surgery~\citep{collins2020augmented, zhang2020real}. Monocular depth estimation is crucial for this reconstruction and has broader applications in surgical robotics and navigation~\citep{xu2023information}. Achieving precise depth estimation in complex surgical environments remains challenging due to low lighting and sparse textures. Deep learning methods have been widely proposed for depth estimation in natural environments. Obtaining accurate ground truth depth for training is difficult due to security, privacy, and professionalism, leading to a focus on self-supervised learning (SSL) techniques where depth estimation is guided by the relationship between video frames~\citep{arampatzakis2023monocular, ozyoruk2021endoslam, wang2024monopcc}. These methods typically involve two networks for depth map and pose estimation and are optimized jointly using photometric loss.

\begin{figure}[t]
\centering
\includegraphics[width=0.95\linewidth]{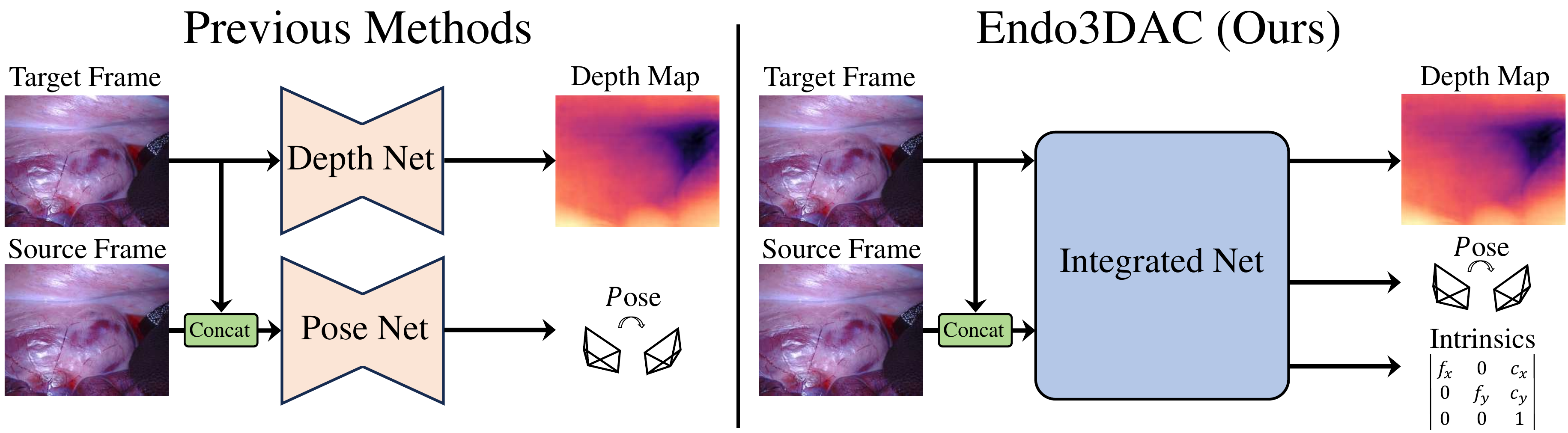}
\caption{Comparison between previous self-supervised depth estimation methods and our proposed Endo3DAC. Previous methods (\textbf{left}) utilize two separate networks to estimate the depth map and the relative pose, which also requires intrinsic parameters for training. In contrast, our proposed method (\textbf{right}) estimates the depth map, relative pose, and camera intrinsic parameters with one integrated network.}
\label{fig:illustration}
\end{figure}

Recently, foundation models have garnered significant attention due to their exceptional performance across a wide range of tasks~\citep{kirillov2023segment, oquab2023dinov2}. By harnessing vast data and advanced training methods, foundation models gain broad knowledge, excelling in tasks involving vision, text, and multi-modal inputs. Nonetheless, foundation models might suffer notable performance deterioration in endoscopic scenes due to large domain gaps~\citep{wang2023sam}. Developing a medical-specific foundational model from scratch presents numerous obstacles, given the limited annotated data and insufficient computational resources. Consequently, there has been more discussion on adapting foundational models for different sub-domains to suit specific application contexts, maximizing the utilization of pre-trained models~\citep{cui2024endodac}.

Most previous SSL depth estimation methods are based on the conventional two-networks strategy, as shown at the left of Fig.~\ref{fig:illustration}. The Depth-Net takes one image as input and outputs the depth map, while Pose-Net takes two images as input and outputs the relative poses. Depth-Net and Pose-Net are parameterized by different sets of parameters and don't share weights. This will result in redundant storage resources and a cumbersome downstream application process. Meanwhile, most researchers made efforts to utilize depth foundation models for only depth estimation tasks but neglected their potential for pose estimation tasks. Besides, such SSL methods require camera intrinsic parameters, which are not always available for every video, to warp the source image to the target image to construct photometric loss. Most current surgical 3D scene reconstruction methods require ground truth in some form indicating a lack of 3D scene reconstruction methods that are highly compatible and can be used with any endoscopic video.

To address these challenges, we first develop an adaptation strategy to efficiently adapt depth estimation foundation models to the surgical domain in an SSL way. We develop Gated Dynamic Vector-Based Low-Rank Adaptation (GDV-LoRA) to customize the foundational model for the surgical domain. GDV-LoRA enables one network to input either one or two images and outputs a depth map or relative pose with intrinsic accordingly. Depth map, relative pose, and camera intrinsic parameters are estimated by different lightweight decoders based on the same encoder. Our framework can be trained only on monocular surgical videos with one network from any camera, regardless of any ground truth, ensuring minimal training expenses. We also introduce a novel SSL loss function to ensure the depth estimation to be scale-invariant and shift-invariant. We further introduce a novel pipeline for dense 3D scene reconstruction using the depth maps, poses, and camera intrinsic parameters predicted by our depth estimation framework by optimizing a sparse set of parameters. The major contributions of this article can be summarized as follows:
\begin{itemize}
    \item[1)] We propose a unified self-supervised adaptation strategy where the depth, poses, and camera's intrinsic parameters are predicted with an integrated network and trained in parallel. Our method only requires surgical videos from any unknown camera to be adapted.
    
    \item[2)] We further present a 3D scene reconstruction method with the outputs generated by our integrated network. The depth maps are optimized to proper scale and shift with a patched global-local rectification process. 
    
    \item[3)] Extensive experiments on four publicly available datasets have demonstrated the superior performance of our proposed method. To the best of our knowledge, we are the first to propose such a unified network and surgical scene reconstruction method where only surgical videos are required for both training and evaluation.

\end{itemize}

This work substantially extends our preliminary work~\citep{cui2024endodac} on adaptation strategy for surgical 3D scene reconstruction at MICCAI'24: (\romannumeral 1) \textbf{Unified Network for Scale and Shift Invariant Depth}: We proposed GDV-LoRA to train a unified network to estimate depth map, relative pose, and camera intrinsic parameters. We also develop a loss function to enforce the depth estimation to be scale and shift invariant in an SSL way. (\romannumeral 2) \textbf{Advancement in 3D Scene Reconstruction}: We further present a 3D scene reconstruction method with the depth maps, poses, and camera intrinsic parameters generated by our integrated network. The depth maps undergo optimization for scale and shift using a global-local rectification process while the poses are concurrently optimized. (\romannumeral 3) \textbf{Comprehensive Evaluation}: We conduct extensive experiments on two more datasets to enhance the comprehensive evaluation.

\section{Related Works}

\subsection{Self-Supervised Depth Estimation }
Self-Supervised Depth Estimation methods for natural scenes have been studied in recent years for dynamic objects, assistance from external sources, and advanced network architectures~\citep{sun2021unsupervised, wang2022unsupervised, han2023self, zhang2023lite, lyu2021hr, bello2024self, lou2024ws}. Most previous Surgical SSL depth estimation methods have been focused on enhancing the reliability of photometric constraints under fluctuated brightness~\citep{wang2024monopcc, shao2022self, rodriguez2023lightdepth, he2024monolot, mahmood2018deep}. Endo-SfMLearner~\citep{ozyoruk2021endoslam} linearly aligned the brightness of the warped image with the mean values and standard deviation of the target image. AF-SfMLearner~\citep{shao2022self} proposed an appearance flow network to estimate the pixel-wise brightness changes of an image. MonoPCC~\citep{wang2024monopcc} is proposed where the photometric constraint is reshaped into a cycle form instead of only warping the source image. The target image ultimately undergoes a cycle-warping process with an image derived from itself, ensuring the constraint remains unaffected by variations in brightness. Many researchers also put efforts into improving the model architecture's efficiency or capability to enhance accuracy. Li et al.~\citep{li2020unsupervised} model the time information by incorporating the LSTM module into the pose network to enhance the precision of pose estimation. Yang et al.~\citep{yang2024self} propose a lightweight network with a tight coupling of convolutional neural networks (CNN) and Transformers for depth estimation. Budd et al.~\citep{budd2024transferring} use optical flow to align depth maps between different views to maintain consistency. However, most current self-supervised learning methods for monocular depth estimation count on the photometric reprojection constraint between adjacent images by training two separate networks (depth and pose networks,) which is tedious for downstream implementation. 

\subsection{Foundation Models for Depth Estimation }
It was not until recent years that foundation models for depth estimation started to reveal their huge potential. MiDaS~\citep{ranftl2020towards} proposes to mix datasets with different annotations while training by aligning the scale and shift of depth maps, which greatly improves the generalization ability of the model across different scenes. They also designed DPT~\citep{ranftl2021vision} to incorporate Vision Transformers~\citep{han2022survey} (ViTs) into dense prediction problems. DINOv2~\citep{oquab2023dinov2} serves as a vision feature foundation model that trains an encoder that outputs a feature vector with well-arranged semantic meanings. They then fine-tuned their model with separate lightweight decoders for many vision tasks, including depth estimation and obtained outstanding performances. Depth Anything~\citep{yang2024depth} trained a depth estimation foundation model with larger-scale labeled and unlabeled datasets. A powerful teacher model is implemented to generate pseudo-labels for supervision. They also developed an auxiliary supervision to mandate the model to inherit rich semantic priors from pre-trained DINOv2~\citep{oquab2023dinov2} encoder. As highlighted earlier, creating a medical-specific foundational model from the ground up poses difficulties because of the limited availability of annotated data in the medical field and insufficient access to computational resources. Consequently, adapting foundational models for particular subdomains emerges as a more effective and proficient approach, albeit one that remains largely unexplored in the realm of surgical depth estimation~\citep{cui2024surgical}.

\subsection{Surgical Scene Reconstruction }
Among existing works, surgical scene reconstruction has achieved success in various directions. Many algorithms~\citep{mahmoud2017slam, mahmoud2018live} are built based on SLAM to deal with illumination changes and scarce texture. Liu et al.~\citep{liu2022sage} develop SAGE in which appearance and geometry prior are exploited to facilitate the SLAM system. Song et al.~\citep{song2018mis} designed a SLAM system to confront deformations during surgical operations. Another direction is based on depth and pose estimations. E-DSSR~\citep{long2021dssr} uses a transformer-based stereoscopic depth perception for efficient depth estimation and a lightweight tool for segmentation to handle tool occlusion for subsequent reconstruction. Wei et al.~\citep{wei2022stereo} propose a depth estimation network robust to texture-less and variant soft tissues. They use surfels to represent the scene and reconstruct the scene with estimated laparoscope poses. Neural Radiance Fields (NeRF)~\citep{mildenhall2021nerf} based methods have been developed quickly in recent years for their amazing performance in novel scene synthesis~\citep{shen2022nerp}. NeRF implicitly uses Multi-Layer Perceptrons (MLP) to represent the volume density and color for any 3D space location. EndoNeRF~\citep{wang2022neural} pioneers the application of NeRF in endoscopic environments through a dual neural fields strategy for capturing tissue deformation and canonical density. Additionally, EndoSurf~\citep{zha2023endosurf} utilizes signed distance functions to represent tissue surfaces, enforcing explicit self-consistency constraints on the neural field. Many other algorithms~\citep{huang2024endo, yang2024efficient, wang2024endogslam} are proposed based on NeRF for the surgical domain and achieve promising results. However, most existing methods require ground truth in some form, e.g., poses or depth maps, besides surgical videos for training or evaluation limiting the general applicability.

\section{Method}
In this section, details of the proposed Endo3DAC framework are presented. First, we illustrate the efficient adaptation strategy for the self-supervised depth estimation framework. Then, the dense scene reconstruction method is provided.

\subsection{Preliminaries}

\subsubsection{Self-supervised Learning}

Formally, let $ D_{t} \in \mathbb{R}^{H\times W}$ be the depth map estimation of target image $ I_{t}$, given $D_{t}$, pose $ T_{t\rightarrow s} \in \mathbb{R}^{4\times 4}$ and camera intrinsic parameters $ K \in \mathbb{R}^{3\times 3}$, a warpping can be performed to construct the warpped image $ I_{s\rightarrow t}$ based on a pixel-to-pixel matching with:
\begin{equation}
p_s=K T_{t \rightarrow s} D_t\left(p_{s \rightarrow t}\right) K^{-1} p_{s \rightarrow t} \text {, }
\label{eq:warp}
\end{equation}

where $p_s$ and $p_{s \rightarrow t}$ denotes the pixel’s homogeneous coordinates in $ I_{s}$ and $ I_{s\rightarrow t}$, respectively, $D(p)$ represents the depth value at position $p$. The warped image from source to target can be obtained by bilinear sampling~\cite{jaderberg2015spatial} as:
\begin{equation}
I_{s \rightarrow t}\left(p_{s \rightarrow t}\right)= \mathbf{BilinearSampler}\left(I_s\left(p_s\right)\right) \text {. }
\label{eq:i}
\end{equation}

The self-supervised photometric loss is defined by:
\begin{equation}
\mathcal{L}_p=\alpha \frac{1-\operatorname{SSIM}\left(I_t, I_{s \rightarrow t}\right)}{2}+(1-\alpha)\left|I_t-I_{s \rightarrow t}\right|,
\end{equation}
combing $\mathcal{L}_1$ loss and structural similarities (SSIM)~\citep{wang2004image} to assess the image difference.

\subsubsection{Low-Rank Adaptation (LoRA)}

LoRA~\cite{hu2021lora} was introduced to adapt foundation models to specific tasks. By integrating trainable rank decomposition matrices into each layer of a network, LoRA substantially decreases the trainable parameters for subsequent tasks while maintaining the frozen pre-trained model weights. To be specific, for a pre-trained weight matrix $W_{0}\in \mathbb{R}^{d \times k}$, LoRA modifies the update to:
\begin{equation}
h = W_{0}x + \Delta Wx = W_{0}x + BAx.
\end{equation}
where $B\in \mathbb{R}^{d \times r}, A\in \mathbb{R}^{r \times k}$ with the rank $r\ll min(d,k)$; $ W_{0}$ is frozen during training and only $A$ and $B$ receive gradient updates.

\subsection{Adaptation Strategy for Self-Supervised Depth Estimation}

\subsubsection{Overview}

\begin{figure*}[t]
\centering
\includegraphics[width=1.0\linewidth]{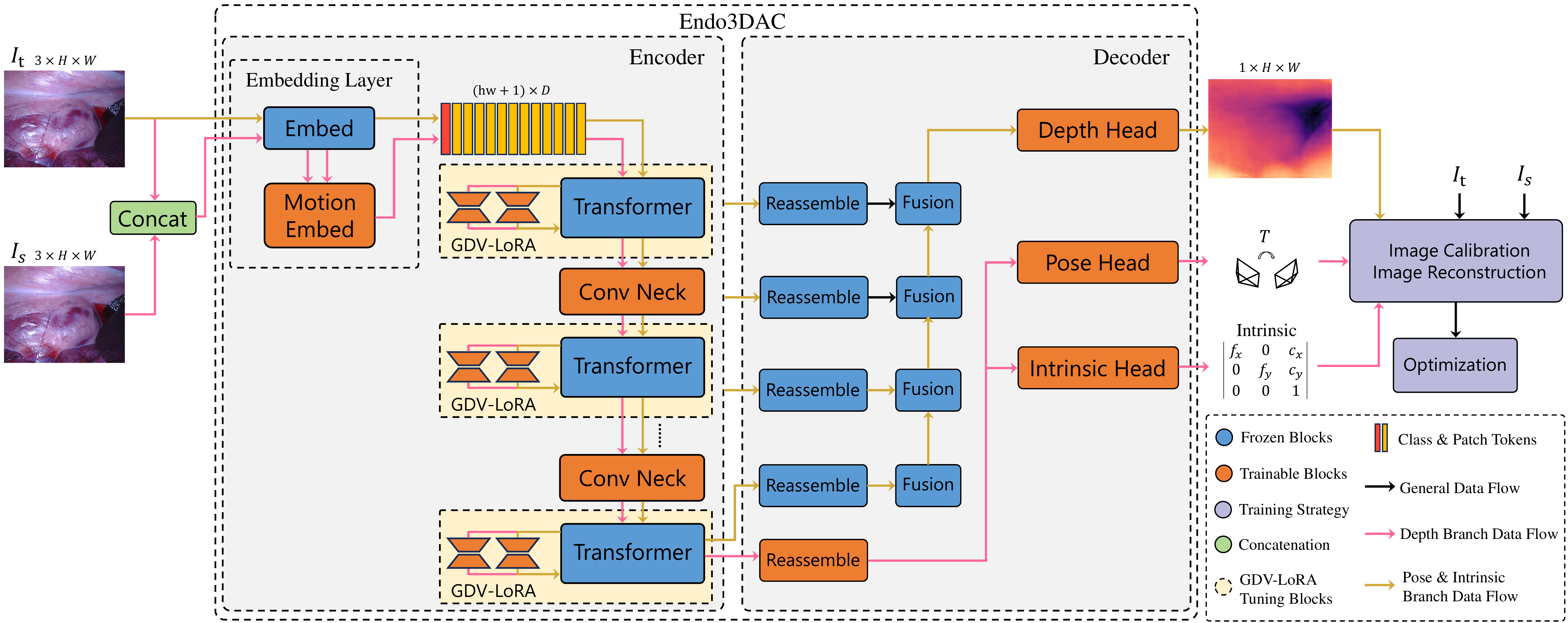}
\caption{Illustration of the proposed Endo3DAC SSL depth estimation framework. ViT-based encoder and DPT-liked decoder pre-trained from Depth Anything~\citep{yang2024depth} are employed. We proposed Gated Dynamic Vector-Based Low-Rank Adaptation (GDV-LoRA) to fine-tune one model for different tasks with different sets of parameters. Convolution neck blocks (Conv Neck) are implemented to enhance the network. Only a few of the parameters are trainable (orange) and separate decoder heads are used to predict depth maps, relative poses, and Intrinsics within one network.}
\label{fig:depthframework}
\end{figure*}

The proposed Endo3DAC aims to adapt the depth estimation foundation model - Depth Anything (DA)~\citep{yang2024depth} - to the endoscopic domain to estimate depth maps, poses, and camera intrinsic parameters with one network in a self-supervised manner. To be specific, Let $ I_{t} \in \mathbb{R}^{3\times H\times W}$ and $ I_{c} \in \mathbb{R}^{6\times H\times W}$ represent a single target image and a concatenation of target image $I_{t}$ and source image $I_{s}$, Endo3DAC outputs depth maps $ D_{t} \in \mathbb{R}^{H\times W}$ when the inputs is $I_{t}$ and outputs poses $ T_{t\rightarrow s} \in \mathbb{R}^{4\times 4}$ with camera intrinsic parameters $ K \in \mathbb{R}^{3\times 3}$ when the inputs is $ I_{c}$. 

Fig.~\ref{fig:depthframework} illustrates the proposed Endo3DAC self-supervised depth estimation framework. We use a ViT-based~\citep{dosovitskiy2020image} encoder and a DPT~\citep{ranftl2021vision} liked decoder as our backbone with pre-trained weights from DA. Instead of fine-tuning the whole network, we implement trainable GDV-LoRA layers, Convolutional Neck blocks, and separate trainable decoder heads with the frozen transformer blocks to efficiently fine-tune the model. 

When the input is a single monocular image, the processing follows the depth branch (yellow) in Fig.~\ref{fig:depthframework}. Formally, $ I_{t} \in \mathbb{R}^{3\times H\times W}$ is first transformed into class and patch tokens $t^{0}\in \mathbb{R}^{(1+hw)\times C}$. The tokens then go through a mixture of $L$ transformer layers with GDV-LoRA adaptation and $J$ convolutional neck blocks to transformed into new representations $t^{l}$, where $h = \frac{H}{p}$, $w = \frac{W}{p}$, $p$ is the size of patches, $C$ is the patches' dimension and $t^{l}$ denotes the outputs of $l$-th transformer layer. Then, a feature representation is obtained by reassembling and an incremental fusion process. Finally, a depth decoder head generates the estimated depth map $ D_{t} \in \mathbb{R}^{H\times W}$.

When the input is $I_{c}$, the processing follows the pose \& intrinsic branch (pink) in Fig.~\ref{fig:depthframework}. The concatenated image is first split into two images, each patchified and concatenated, followed by linear projections to project the dimensions from $2C$ to $C$. Tokens then go through the same process as before where the final feature representation is reassembled and predicts a relative pose matrix $ T \in \mathbb{R}^{4\times 4}$ with pose decoder head and camera intrinsic parameters $K \in \mathbb{R}^{3\times 3}$ with intrinsic decoder head. The depth estimation $D_{t}$ is reprojected back onto the 2-D plane with $T$ and $K$ to create the reconstructed image. The model can, therefore, be optimized by minimizing the loss between the reconstructed image and the target image.

\subsubsection{Foundation Model Adaptation Strategy}

\begin{figure}[t]
\centering
\includegraphics[width=0.85\linewidth]{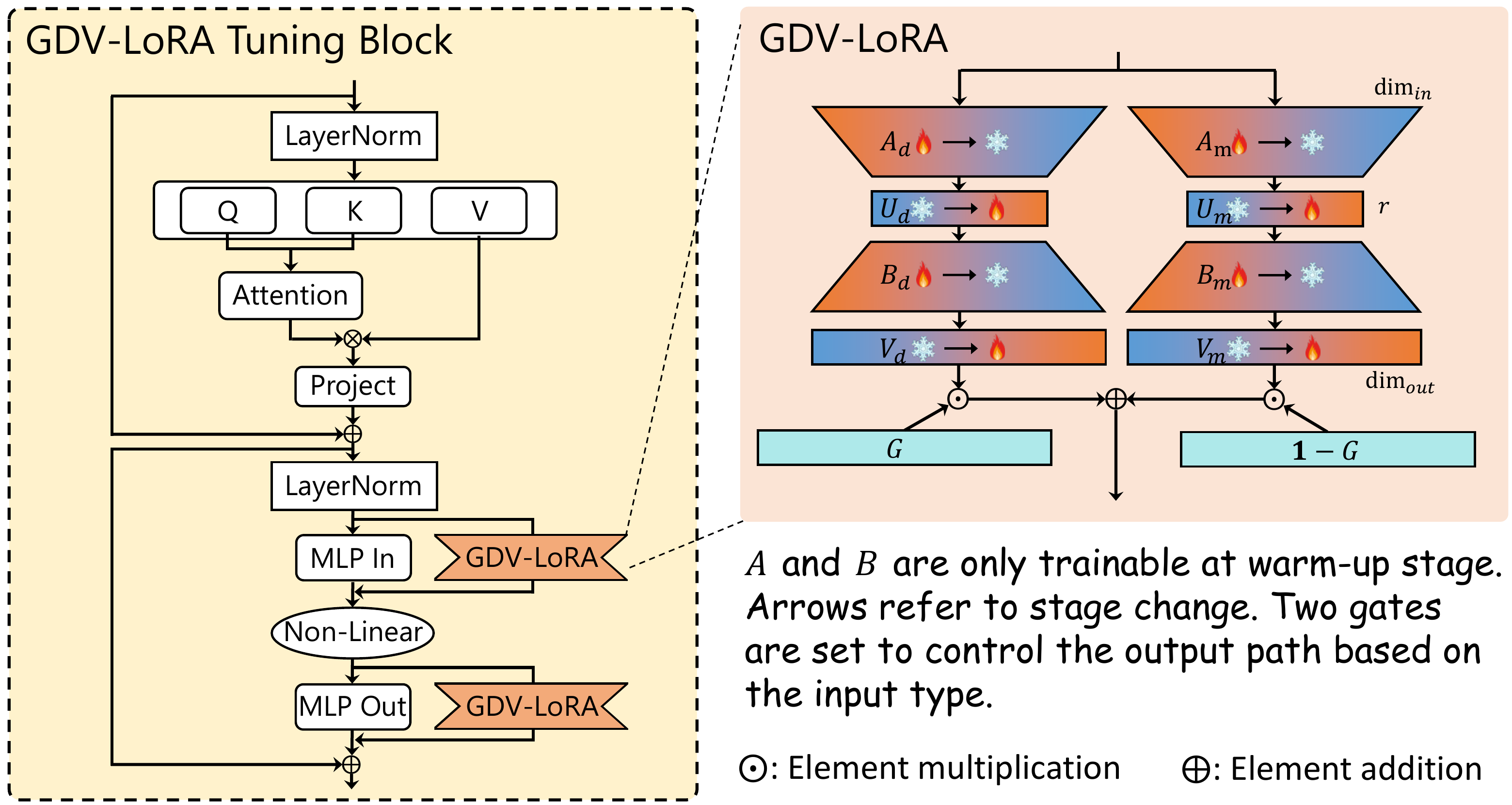}
\caption{Illustration of GDV-LoRA Tuning Block. Different sets of parameters are used for the depth estimation task and pose-intrinsic estimation task with a control gate. We use the gradient color and arrows to represent the dynamic variation between training and frozen states.}
\label{fig:gdvlora}
\end{figure}

The inherent visual features within a depth estimation foundation model could also benefit pose estimation because the direction and magnitude of depth variation can, to some extent, reflect changes in poses. Therefore, we innovatively introduce Gated Dynamic Vector-Based Low-Rank Adaptation (GDV-LoRA) to fine-tune one model for different tasks more efficiently. GDV-LoRA is applied exclusively to two MLP layers for thorough adaptation, as shown in Fig.~\ref{fig:gdvlora}. Specifically, GDV-LoRA updates the pre-trained weight matrix $W_{0}\in \mathbb{R}^{d \times k}$ to:
\begin{equation}
\begin{aligned}
x_{out} = & W_{0} x_{in}+ G \odot \Lambda_{vd} B_d \Lambda_{ud} A_d x_{in} \\ &+ (\textbf{1}-G) \odot \Lambda_{vm} B_m \Lambda_{um} A_m x_{in}, \\
\end{aligned}
\end{equation}
where $x_{in}, x_{out}$ are inputs and outputs of MLP layers; $B_d, B_m\in \mathbb{R}^{d \times r}$, $A_d, A_m\in \mathbb{R}^{r \times k}$ are trainable LoRA layers; $\Lambda_{vd}, \Lambda_{vm}\in \mathbb{R}^{k \times k}$ and $\Lambda_{ud}, \Lambda_{um}\in \mathbb{R}^{r \times r}$ are trainable vectors $V_d$, $V_m$, $U_d$ and $U_m$ in diagonal matrices form; $G$ is the gated vector, $\textbf{1}$ is all-ones matrix with same size of $G$; $\odot$ denotes element-wise multiplication. $G = 0$ if $C_{in}=3$ and $G = 1$ if $C_{in}=6$, where $C_{in}$ is the channel value of the input image. GDV-LoRA thus enables fine-tuning different sets of parameters for different tasks within the same frozen pre-trained network. Only LoRA layers $A$, $B$ receive gradient updates while $U$, $V$ are frozen at the warm-up stage. The state of GDV-LoRA shifts dynamically where $A$, $B$ are frozen and $U$, $V$ become trainable after a warm-up phase. This approach allows us to refine the model using a solid initialization, focusing on training the model with fewer parameters. 

As demonstrated by~\citep{park2021vision}, Vision Transformers (ViTs) tend to diminish high-frequency signals, potentially impacting depth estimation adversely. Thus, drawing inspiration from~\citep{yao2024vitmatte}, we utilize convolution neck blocks to improve our approach. We incorporate a convolutional neck block which consists of three convolutional layers with LayerNorm and a residual connection after each $3^{rd}, 6^{th}, 9^{th}$ and $12^{th}$ transformer GDV-LoRA tuning block.

\subsubsection{Self-supervised Scale- and Shift-Invariant Loss}
Scale- and shift-invariant loss were proposed for supervised depth estimation to fully utilize different datasets from diverse scenes with different scales and shifts~\citep{Ranftl2022}. We design a self-supervised scale- and shift-invariant loss for our framework to enforce the depth estimation to be affine-invariant without the supervision of ground truth. We aim to align the depth estimation of the same area from different views. Unlike~\citep{budd2024transferring} which utilizes optical flow to align different views for different models, we directly use relative poses between different views to maintain consistency for one network. Formally, let $ D_{s} \in \mathbb{R}^{H\times W}$ and $ D_{t} \in \mathbb{R}^{H\times W}$ be the depth map estimation of source image $ I_{s}$ and target image $ I_{t}$. With equation~\ref{eq:warp}, instead of warping the color image, we warp the depth map from the source to the target image to obtain $ D_{s\rightarrow t}$ with:
\begin{equation}
D_{s \rightarrow t}\left(p_{s \rightarrow t}\right)= \mathbf{BilinearSampler}\left(D_s\left(p_s\right)\right) \text {. }
\label{eq:d}
\end{equation}

We define self-supervised scale- and shift-invariant loss as:
\begin{equation}
\mathcal{L}_{s-ssi}=\frac{1}{H W} \sum_{i=1}^{H W} \rho\left(D_{t}(i)), D_{s \rightarrow t}(i))\right)
\end{equation}
where $D(i)$ refers to the $i$-th value of depth map $D$, $\rho$ is the mean absolute error loss: $\rho\left(D_{t}(i)), D_{s \rightarrow t}(i))\right) = |\hat{D_{t}}(i) - \hat{D_{s \rightarrow t}}(i)|$ where $\hat{D_{t}}(i)$ and $\hat{D_{s \rightarrow t}}(i)$ are the scale and shifted versions of depth map: $\hat{D}(i) = \frac{D(i) - t(D)}{s(D)}$. $t(D)$ and $s(D)$ are functions used to adjust the depth map to have zero translation and unit scale:
\begin{equation}
t(D) = median(D), s(D)=\frac{1}{H W} \sum_{i=1}^{H W}\left|D(i)-t(D)\right|
\end{equation}

The final loss function combines the above-mentioned losses with an edge-regularization loss $\mathcal{L}_e$~\citep{godard2019digging} expressed as: 
\begin{equation}
\mathcal{L}_{d} = \lambda_{p}\mathcal{L}_p + \lambda_{s-ssi}\mathcal{L}_{s-ssi} + \lambda_{e}\mathcal{L}_e, 
\end{equation}
where $\lambda_{p}, \lambda_{e}$ and $\lambda_{s-ssi}$ are weights of losses.

\subsection{Dense Scene Reconstruction}

\subsubsection{Overview}

\begin{figure*}[t]
\centering
\includegraphics[width=1.0\linewidth]{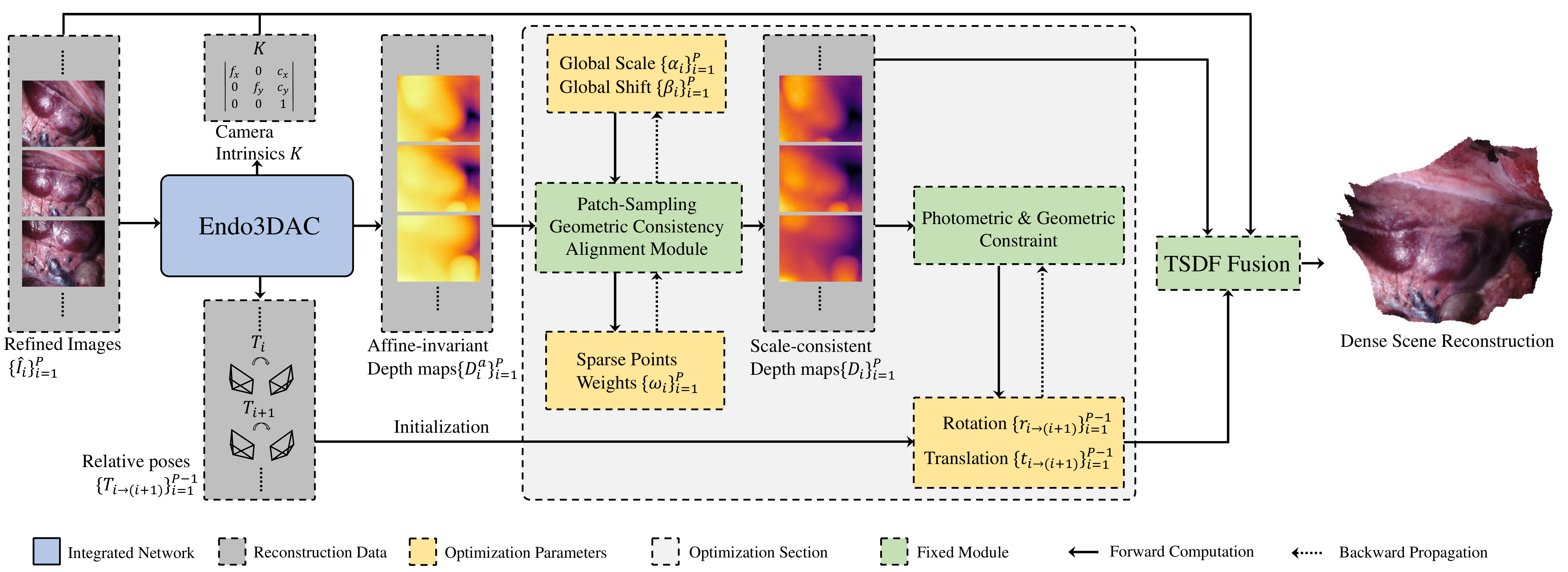}
\caption{The proposed dense scene reconstruction framework. Given a monocular surgical video, we first use Endo3DAC to generate the depth maps, relative poses, and camera intrinsic parameters. Then, we propose a patch-sampling geometric consistency alignment module to optimize a small number of variables to align the scale and shift among all depth maps. Poses are initialized with estimated poses and optimized concurrently. Finally, we obtain a dense scene reconstruction with the optimized depth maps and relative poses.}
\label{fig:recon}
\end{figure*}

Estimated depth maps and poses from Endo3DAC might be inconsistent. Direct transformation to point clouds will be inaccurate and noisy. We propose a lightweight pipeline to optimize depth maps jointly and camera poses with only monocular surgical videos for dense scene reconstruction, as shown in Fig.~\ref{fig:recon}. 

With sampled images $\left\{ I_{i}\right\}_{i=1}^{P}$, we first utilize the appearance flow network from the previous section to calibrate for the inconsistent lighting to obtain refined Images $ \left\{ \hat{I}_{i}\right\}_{i=1}^{P}$. Then we use Endo3DAC to obtain the depth maps $\left\{ D_{i}^{a}\right\}_{i=1}^{P}$, relative poses $\left\{ T_{i\to i+1}\right\}_{i=1}^{P-1}$ and camera intrinsics $K$. Inspired by~\citep{xu2023frozenrecon}, we propose a patch-sampling geometric consistency alignment module $F(\cdot ,\cdot ,\cdot ,\cdot )$ to recover unknown scale and shift of depth maps and obtain affine-consistent depth $D_{i}$:
\begin{equation}
D_{i} = F(D_{i}^{a} ,\alpha_{i} ,\beta_{i} ,w_{i} ),
\end{equation}

where $\alpha_{i}$, $\beta_{i}$ and $w_{i}$ are the variables to be optimized. We also optimize the rotation $\left\{ r_{i\to i+1}\right\}_{i=1}^{P-1}$and translation $\left\{ t_{i\to i+1}\right\}_{i=1}^{P-1}$ between adjacent refined images which is initialized by $\left\{ T_{i\to i+1}\right\}_{i=1}^{P-1}$. The optimization is supervised by the color and depth difference between warped points and target points to ensure multi-frame consistency. Ultimately, we can construct the dense 3D scene reconstruction by TSDF fusion~\citep{zeng20173dmatch} with the camera intrinsics parameters, optimized affine-consistent depth maps, and camera poses.

\subsubsection{Optimization Process}
The estimated depth maps $\left\{ D_{i}^{a}\right\}_{i=1}^{P}$ are affine-invariant but with unknown scale and shift, which will lead to distorted points clouds. Therefore, we first perform a geometric consistency alignment module~\citep{xu2023frozenrecon} $F(\cdot)$ consisting of a global and local alignment to align depth maps. Global scale $\alpha_{i}$ and shift $\beta_{i}$ are first utilized to construct the globally aligned depth map $D_{i}^{g}$ with following equation:
\begin{equation}
D_{i}^{g} = \alpha_{i}D_{i}^{a} + \beta_{i}.
\end{equation}

Then, we perform the local alignment by optimizing a subset of the globally aligned depth map. To be specific, we first segment $D_{i}^{g}\in \mathbb{R}^{H\times W}$ input patches with resolution of $(P,P)$. $N$ patches are obtained and we uniformly sample one point within each patch where $N = HW/P^2$. This patch-sampling strategy is more robust to local optimum problems by ensuring proper distance among all points. We then compute the sampled anchor points $\left\{ w_{i,j} \cdot D_{i}^{g}(p_{j})\right\}_{j=1}^{N}$ by multiplying sampled depth $ D_{i}^{g}(p_{j})$ with weights $w_{i,j}$, where $\left\{p_j\right\}_{j=1}^{N}$ denotes the sampled points. The local scale map $A_{i}\in \mathbb{R}^{H\times W}$ and local shift map $B_{i}\in \mathbb{R}^{H\times W}$ are then computed with $D_{i}^{g}$ and $\left\{ w_{i,j} \cdot D_{i}^{g}(p_{j})\right\}_{j=1}^{N}$ by locally weighted linear regression method~\citep{xu2024toward}. Finally, we achieve the affine-consistent depth map with:
\begin{equation}
D_{i} = A_{i} \odot D_{i}^{g} + B_{i}.
\end{equation}
 
We initialize $\left\{ r_{i\to i+1}\right\}_{i=1}^{P-1}$ and $\left\{ t_{i\to i+1}\right\}_{i=1}^{P-1}$ with the predicted poses from Endo3DAC network and transform them to pose matrices $\left\{ T_{i\to i+1}\right\}_{i=1}^{P-1}$.

We implement both pixel-to-pixel photometric and geometric constraints described by equation~\ref{eq:warp}, ~\ref{eq:i} and ~\ref{eq:d} for color and depth consistency in scene reconstruction to optimize the proposed variables $\left\{ \alpha_{i}\right\}_{i=1}^{P}$, $\left\{ \beta_{i}\right\}_{i=1}^{P}$, $\left\{ w_{i}\right\}_{i=1}^{P}$, $\left\{ r_{i\to i+1}\right\}_{i=1}^{P-1}$and $\left\{ t_{i\to i+1}\right\}_{i=1}^{P-1}$ described below:
\begin{equation}
\mathcal{L}_{p c}=\frac{1}{|V|} \sum_{\substack{p \in V \\(i, j) \in K}}\left|I_{i}(p_{i})-I_{j}(p_{i \to j})\right|,
\end{equation}
\begin{equation}
\mathcal{L}_{g c}=\frac{1}{|V|} \sum_{\substack{\mathbf{p} \in V \\(i, j) \in K}} \frac{\left|D_{j}(p_{i \to j})-D_{i \to j}(p_{i})\right|}{D_{j}(p_{i \to j})+D_{i \to j}(p_{i})},
\end{equation}
where $V$ denotes the set of valid points projected from frame $i$ to frame $j$, $K$ is the selected keyframe set. The final optimization constraint is defined by:
\begin{equation}
\mathcal{L}_{r} = \lambda_{pc}\mathcal{L}_{pc} + \lambda_{gc}\mathcal{L}_{gc} + \lambda_{regu}\mathcal{L}_{regu}, 
\end{equation}
where $\mathcal{L}_{regu}$ is a regularization constraint, $\lambda_{pc}, \lambda_{gc}$ and $\lambda_{regu}$ are weights of losses. All the keyframes are selected with a local-global sampling method following~\citep{xu2023frozenrecon}.

\section{Experiments}

\subsection{Dataset}
\subsubsection{SCARED Dataset}
Initially introduced for a competition at MICCAI 2019, SCARED~\citep{allan2021stereo} comprises 35 endoscopic videos capturing 22,950 frames showcasing the abdominal anatomy of fresh porcine cadavers, recorded using a da Vinci Xi endoscope. Each video is paired with ground truth depth maps obtained via a projector, alongside ground truth poses and camera intrinsic details. We adopted the dataset partitioning strategy outlined in ~\citep{shao2022self, yang2024depth}, dividing the SCARED dataset into 15,351, 1,705, and 551 frames for the training, validation, and test sets, respectively.

\subsubsection{SimCol Dataset}
SimCol~\citep{rau2023bimodal} is from MICCAI 2022 EndoVis challenge. It comprises more than 36,000 colonoscopy images with depth annotations with size of 475 × 475. We adhere to the guidelines provided on their official website, dividing the dataset into 28,776 frames for training and 9,009 frames for testing.

\subsubsection{Hamlyn Dataset}
Hamlyn Dataset~\citep{mountney2010three} is a large public endoscopic dataset that includes demanding sequences capturing intracorporeal scenes characterized by subtle textures, deformations, reflections, surgical instruments, and occlusions. The ground truth depth maps are obtained by~\citep{recasens2021endo} with the stereo matching software Libelas. We choose 21 videos for testing followed~\citep{recasens2021endo}.

\subsubsection{C3VD Dataset}
Colonoscopy 3D Video Dataset (C3VD)~\citep{bobrow2023} is a public dataset acquired with a high-definition clinical colonoscope and high-fidelity colon models. It provided 22 colonic videos with ground truth depth maps, surface normal maps, optical flow maps, occlusion maps, poses, and complete 3D models. We followed previous works~\citep{rodriguez2023lightdepth, paruchuri2024leveraging} and selected 8 videos for testing.

\subsection{Experiments Setting}

\subsubsection{Implementation Details}
The framework is implemented with PyTorch on NVIDIA RTX 4090 GPU. We utilize AdamW~\citep{loshchilov2017decoupled} optimizers for Endo3DAC and the scene reconstruction pipeline with initial learning rates of $1 \times 10^{-4}$. We utilize the ViT-Base model pre-trained from Depth Anything~\citep{yang2024depth} as our fine-tuning model. For the Endo3DAC SSL depth estimation framework, The rank for GDV-LoRA is set to 4, warm up step is set to 5000 and the weights $\alpha$, $\mathcal{L}_{p}$, $\mathcal{L}_{c}$ and $\mathcal{L}_{s-ssi}$ are set to $0.85$, $1$, $0.1$ and $0.01$ respectively. Batch size is set to 8 with 20 epochs in total. For the dense scene reconstruction pipeline, we optimize the parameters for 3 epochs, each with 1000 iterations. $\mathcal{L}_{pc}$, $\mathcal{L}_{gc}$ and $\mathcal{L}_{regu}$ are set to $2$, $0.5$, $0.01$ respectively.

\subsubsection{Evaluation Metrics}
We employ 5 standard metrics for depth evaluation, which are Abs Rel, Sq Rel, RMSE, RMSE log, and $\delta$, following previous works~\citep{wang2024monopcc, yang2024depth, shao2022self, yang2024self, cui2024surgical}. We align the predicted depth with ground truth depth before evaluation as in~\citep{yang2024depth}. A maximum value is set for each dataset to cap the depth maps which are 150mm, 200mm, 300mm, and 100mm for SCARED, SimCol3D, Hamlyn, and C3VD, respectively.

For the pose evaluation, we perform a 5-frame pose evaluation following~\citep{wang2024monopcc, shao2022self, yang2024self} and adopt the metric of absolute trajectory error (ATE) and relative pose error (RPE).

Furthermore, we utilize 3D metrics including Accuracy (Acc), Completeness (Comp), Chamfer distance (Cham), Precision (Prec), Recall (Rec), and F1-score (F1) to evaluate our proposed 3d scene reconstruction framework. Reconstructed points cloud and ground truth points cloud are first registered with the ICP~\citep{besl1992method} algorithm before evaluation.

\subsection{Experiments on Depth}

\subsubsection{Quantitative Evaluation on SCARED and SimCol}

\begin{table*}[t]
\caption{Quantitative depth comparison on \textbf{SCARED dataset} and \textbf{SimCol dataset} of SOTA self-supervised learning depth estimation methods. The best results are in bold and the second-best results are underlined. "R.I." refers to whether the method requires camera intrinsic parameters.}
\centering
\label{tab:depth_main_SCARED_SimCol}  
\resizebox{1.0\textwidth}{!}{
\begin{tabular}{c|c|ccccc|ccccc}
\toprule
\multirow{2}{*}{Method} & \multirow{2}{*}{R.I.}  & \multicolumn{5}{c|}{SCARED} & \multicolumn{5}{c}{SimCol} \\
\cline{3-12}
 &  & Abs Rel $\downarrow$ & Sq Rel $\downarrow$ & RMSE $\downarrow$ & RMSE log $\downarrow$ & $\delta \uparrow$ & Abs Rel $\downarrow$ & Sq Rel $\downarrow$ & RMSE $\downarrow$ & RMSE log $\downarrow$ & $\delta \uparrow$ \\ \midrule 
Monodepth2~\citep{godard2019digging} & \cmark & 0.083 & 0.842 & 6.666 & 0.111 & 0.934 & 0.212 & 0.992 & 1.165 & 0.243 & 0.763 \\
Endo-SfM~\citep{ozyoruk2021endoslam} & \cmark & 0.069 & 0.666 & 6.117 & 0.096 & 0.960 & 0.200 & 0.918 & 1.127 & 0.238 & 0.778 \\
DA~\citep{yang2024depth} & - & 0.073 & 0.691 & 5.909 & 0.095 & 0.958 & 0.273 & 1.471 & 1.583 & 0.266 & 0.732 \\ 
HR-Depth~\citep{lyu2021hr} & \cmark & 0.068 & 0.575 & 5.683 & 0.092 & 0.962 & 0.110 & 0.720 & \underline{0.570} & 0.110 & 0.947 \\
MonoViT~\citep{zhao2022monovit} & \cmark & 0.062 & 0.470 & 5.042 & 0.082 & 0.976 & 0.082 & 0.295 & 0.576 & \underline{0.104} & 0.951 \\
Lite-Mono~\citep{zhang2023lite} & \cmark & 0.057 & 0.453 & 4.967 & 0.079 & 0.975 & 0.133 & 1.375 & 0.606 & 0.107 & 0.954 \\
DA(fine-tuned)~\citep{yang2024depth} & \cmark & 0.058 & 0.452 & 4.885 & 0.102 & 0.974 & 0.089 & 0.421 & 0.599 & 0.112 & 0.948 \\
Af-SfM~\citep{shao2022self} & \cmark & 0.055 & 0.402 & 4.625 & 0.076 & 0.978 & 0.086 & 0.358 & 0.585 & \underline{0.104} & 0.954 \\  \hline

Endo3DAC (Ours) & \xmark &  \underline{0.046} & \underline{0.307} & \underline{4.063} & \underline{0.065} & \underline{0.985} & \underline{0.082} & \underline{0.287} & 0.611 & 0.107 & \underline{0.955} \\
Endo3DAC (Ours) & \cmark & \textbf{0.045} & \textbf{0.290} & \textbf{4.040} & \textbf{0.064} & \textbf{0.987} & \textbf{0.076} & \textbf{0.266}  &   \textbf{0.555} &   \textbf{0.101}  &  \textbf{0.957} \\
\bottomrule

\end{tabular}}
\end{table*}

\begin{table*}[t]
\caption{Zero-shot quantitative depth comparison on \textbf{Hamlyn dataset} and \textbf{C3VD dataset}. The results on the Hamlyn dataset are self-supervised monocular trained on the SCARED dataset and the results on the C3VD dataset are self-supervised monocular trained on the SimCol3D dataset. The best results are in bold and the second-best results are underlined.}
\centering
\label{tab:depth_main_Hamlyn_C3VD}  
\resizebox{1.0\textwidth}{!}{
\begin{tabular}{c|c|ccccc|ccccc}
\toprule
\multirow{2}{*}{Method} & \multirow{2}{*}{R.I.} & \multicolumn{5}{c|}{Hamlyn} & \multicolumn{5}{c}{C3VD} \\ \cline{3-12}
 & & Abs Rel $\downarrow$ & Sq Rel $\downarrow$ & RMSE $\downarrow$ & RMSE log $\downarrow$ & $\delta \uparrow$ & Abs Rel $\downarrow$ & Sq Rel $\downarrow$ & RMSE $\downarrow$ & RMSE log $\downarrow$ & $\delta \uparrow$  \\ \midrule 
Monodepth2~\citep{godard2019digging} & \cmark & 0.197 & 7.135 & 16.459 & 0.229 & 0.735 & 0.170 & 2.317 & 9.276 & 0.225 & 0.769 \\
Endo-SfM~\citep{ozyoruk2021endoslam} & \cmark & 0.210 & 13.235 & 19.511 & 0.234 & 0.755 & 0.164 & 2.232 & 9.311 & 0.217 & 0.770 \\
DA~\citep{yang2024depth} & - & 0.170 & 6.093 & 14.523 & 0.204 & 0.791 & 0.246 & 6.423 & 14.501 & 0.298 & 0.684 \\  
HR-Depth~\citep{lyu2021hr} & \cmark & 0.202 & 11.106 & 19.105 & 0.231 & 0.758 & 0.152 & 2.102 & 9.293 & 0.196 & 0.787 \\
MonoViT~\citep{zhao2022monovit} & \cmark & 0.193 & 10.512 & 18.028 & 0.220 & 0.769 & 0.116 & 1.014 & 6.712 & 0.153 & 0.881 \\
Lite-Mono~\citep{zhang2023lite} & \cmark & 0.179 & 6.366 & 15.196 & 0.216 & 0.754 & 0.111 & \underline{0.762} & \underline{5.474} & 0.155 & \underline{0.890} \\
DA(fine-tuned)~\citep{yang2024depth} & \cmark & 0.180 & 6.217 & 15.146 & 0.211 & 0.770 &  0.121 &  1.155 &  6.239 &  0.165 & 0.874 \\
Af-SfM~\citep{shao2022self} & \cmark & 0.185 & 7.021 & 15.685 & 0.217 & 0.767 &  0.117 &  1.324 &  7.520 &  0.171 & 0.882 \\ \hline

Endo3DAC (Ours) & \xmark & \textbf{0.157} & \textbf{4.910} & \textbf{13.323} & \textbf{0.190} & \textbf{0.795} & \underline{0.105} & 0.905 & 6.180 &       \underline{0.137} &  0.880 \\
Endo3DAC (Ours) & \cmark & \underline{0.166} & \underline{6.000} & \underline{14.256} & \underline{0.198} & \underline{0.793} &   \textbf{0.083}      &       \textbf{0.584}      &       \textbf{4.655}      &       \textbf{0.107}      &       \textbf{0.949} \\
\bottomrule

\end{tabular}}
\end{table*}

We compares our methods with six other state-of-the-art monocular self-supervised depth estimation methods including Monodepth2~\citep{godard2019digging}, Endo-SfM~\citep{ozyoruk2021endoslam}, HR-Depth~\citep{lyu2021hr}, MonoViT~\citep{zhao2022monovit}, Lite-Mono~\citep{zhang2023lite} and Af-SfM~\citep{shao2022self}. We also compare our method with the pre-trained foundation model Depth-Anything (DA)~\citep{yang2024depth} and fine-tune DA on SCARED and SimCol datasets in an SSL way to obtain DA(fine-tuned). We train and evaluate on the SCARED and SimCol datasets separately. The results are shown in Table~\ref{tab:depth_main_SCARED_SimCol} where our method achieves the best performance in all five evaluation metrics on both SCARED and SimCol datasets. All the previous methods require camera intrinsic parameters for training while our framework can train without it and still achieve the best performance on the SCARED dataset and obtain the best Abs Rel, Sq Rel, and $\delta$ on the SimCol dataset compared to other methods.

DA was trained on large-scale datasets of natural scenes but still exceeds Monodepth2 and Endo-SfM in SCARED. However, without training on the SimCol dataset, DA's performance degrades significantly in all metrics, demonstrating the great generalization of foundation models in natural scenes is not reliable in medical scenes. With fine-tuning on the decoder, DA(fine-tuned) improves by a great margin with only 5.62\% and 4.08\% decreases in RMSE and $\delta$ than Af-SfM, a method designed for solving surgical lighting problems. 

\subsubsection{Zero-Shot Evaluation on Hamlyn and C3VD}
To demonstrate the generalization ability of our proposed method, we zero-shot evaluate our method trained on SCARED with Hamlyn, and our method trained on SimCol with C3VD. Table~\ref{tab:depth_main_Hamlyn_C3VD} lists the zero-shot comparison results. As can be seen, a varying degree of degradation occurs on both datasets for all methods, while Endo3DAC is still the best on all the evaluation metrics on both datasets, revealing its strong generalization ability across different organs and cameras. Endo3DAC even results in the best performance on Hamlyn without the knowledge of camera intrinsic. Also, Endo3DAC achieves the only method with $\delta$ greater than 0.9 and significantly surpasses the second-best Lite-Mono by 23.36\% and 14.96\% in Sq Rel and RMSE on C3VD, respectively.

\subsubsection{Model Size and Inference Speed Analysis}

\begin{table*}[t]
\caption{Comparison of the amount of parameters and inference speed of Depth Net and Pose Net.}
\centering
\label{tab:depth_size}  
\resizebox{0.8\textwidth}{!}{
\begin{tabular}{c|c|ccc|ccc|cc}
\toprule
\multirow{2}{*}{Method} & \multirow{2}{*}{Backbone} & \multicolumn{3}{c|}{Total Parameters. (M)} & \multicolumn{3}{c|}{Trainable Parameters. (M)} & \multicolumn{2}{c}{Inference speed. (ms)}\\ \cline{3-10}
 & & Depth-Net & Pose-Net & Overall & Depth-Net & Pose-Net & Overall & Depth-Net & Pose-Net \\ \midrule 
MonoViT~\citep{zhao2022monovit} & ViT & 27.9 & 13.0 & 40.9 &  27.9 & 13.0 & 40.9 & 49.1 & 4.0 \\
HR-Depth~\citep{lyu2021hr} & ResNet-18 & 14.6 & 13.0 & 27.6 &  14.6 & 13.0 & 27.6 & 13.8 & 4.0 \\
Lite-Mono~\citep{zhang2023lite} & Lite-Mono & 3.1 & 13.0 & 16.1 &  3.1 & 13.0 & 16.1 & 4.6 & 4.0 \\
Af-SfM~\citep{shao2022self} & ResNet-18 & 14.8 & 13.0 & 27.8 &  14.8 & 13.0 & 27.8 & 4.0 & 4.0 \\ \midrule 
Endo3DAC (Ours)& ViT & \multicolumn{2}{c|}{107.6} & 107.6 & 1.4 & 8.8 & 10.2 & 12.0 & 11.2 \\

\bottomrule

\end{tabular}}
\end{table*}

Table~\ref{tab:depth_size} presents a comparison of several methods on The amount of parameters and inference speed. Foundation models typically employ large architectures leading our Endo3DAC to have 107.6 million parameters in total. Lite-Mono is a lightweight network that only contains 3.1 million parameters for Depth-Net. But in terms of trainable parameters, Endo3DAC only utilizes 1.4 million and 8.8 million for Depth-Net and Pose-Net, respectively, which are the minimum compared to other methods. Fewer trainable parameters result in less computation resources and less time for fine-tuning which is beneficial for further medical application. Endo3DAC performs slower in inference speeds compared to alternative methods. However, with a speed of 12.0 ms per frame for depth estimation and 11.2 ms per frame for pose estimation, our framework is still able to support real-time implementation, rendering it suitable for a range of real-time surgical applications.

\subsubsection{Qualitative Evaluation}

\begin{figure*}[t]
\centering
\includegraphics[width=0.95\linewidth]{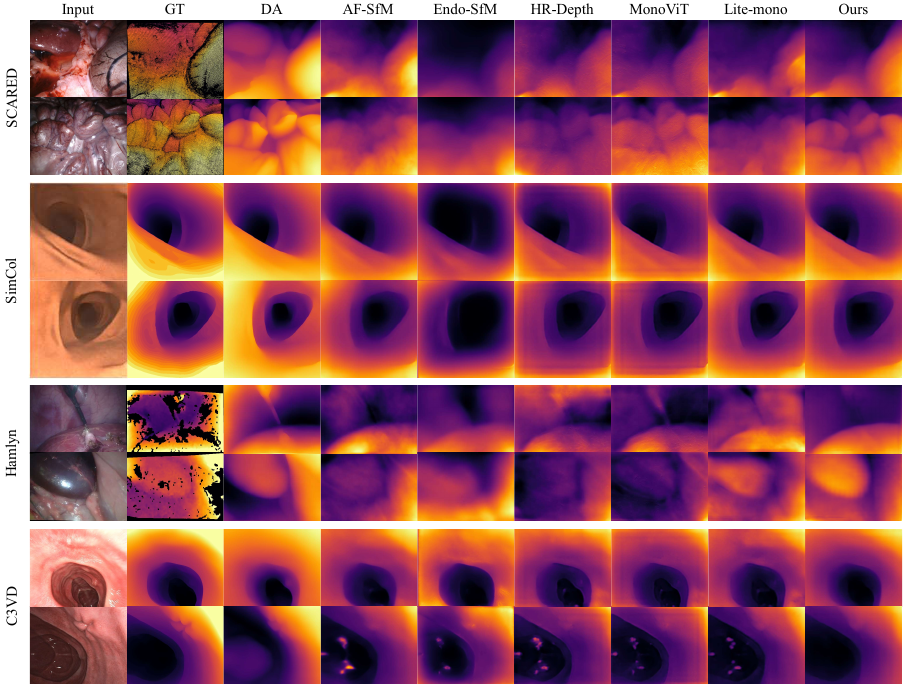}
\caption{Qualitative depth comparison on the SCARED, SimCol, Hamlyn, and C3VD datasets. Our method can generate more continuous and reasonable depth maps with clearer edges, especially for the zero-shot performance on Hamlyn and C3VD, showing the great generalization ability of our method. Endo3DAC generates more reasonable and smoother depth maps on Hamlyn and C3VD.}
\label{fig:depthvis}
\end{figure*}

Fig.~\ref{fig:depthvis} compares the four datasets qualitatively. Our model produces more accurate depth estimation with smoother surfaces and clear edges. The qualitative results also demonstrate the superior generalization ability on unseen datasets. Especially note the last column, all the other methods show erroneous estimation in the left-bottom region affected by the tissue structure and reflection of light while our method maintains smooth and clear estimation.

\subsubsection{Ablation Study}

\begin{table*}[t!]
\caption{Ablation study on the main modules of Endo3DAC. Specifically, we (1) use the original LoRA~\citep{hu2021lora} to replace GDV-LoRA for depth estimation; (2) use the original LoRA~\citep{hu2021lora} to replace GDV-LoRA for pose and intrinsic estimation;(3) remove the Convolution Neck blocks; (4) disable the proposed self-supervised scale and shift invariant loss.}
\fontsize{8}{10}\selectfont
\centering
\resizebox{0.8\textwidth}{!}{
\begin{tabular}{c|c|c|c|ccccc}
\hline 
\makecell[c]{GDV-LoRA \\ (Depth)}  & \makecell[c]{GDV-LoRA \\ (Pose\&Intrinsic)}    & Conv Neck & Self-SSI & Abs Rel $\downarrow$ & Sq Rel $\downarrow$ & RMSE $\downarrow$ & RMSE log $\downarrow$ & $\delta \uparrow$ \\ \hline 
\xmark & \xmark & \xmark & \xmark & 0.058 & 0.412 & 4.737 & 0.080 & 0.973 \\ 
\checkmark & \xmark & \xmark & \xmark & 0.050 & 0.342 & 4.402 & 0.070 & 0.981 \\ 
\xmark & \checkmark & \xmark & \xmark & 0.052 & 0.355 & 4.471 & 0.071 & 0.981 \\ 
\xmark & \xmark & \checkmark & \xmark & 0.055 & 0.367 & 4.490 & 0.076 & 0.977 \\ 
\checkmark & \checkmark & \xmark & \xmark & 0.051 & 0.342 & 4.374 & 0.070 & 0.981 \\  
\checkmark & \xmark & \checkmark & \xmark & 0.052 & 0.352 & 4.432 & 0.075 & 0.979 \\ 
\xmark & \checkmark & \checkmark & \xmark & 0.050 & 0.331 & 4.287 & 0.070 & 0.982 \\ 
\checkmark & \checkmark & \checkmark & \xmark & 0.048 & 0.310 & 4.113 & 0.067 & 0.985 \\ 
\checkmark & \checkmark & \checkmark & \checkmark & 0.046 & 0.307 & 4.063 & 0.065 & 0.985   \\ \hline 

\end{tabular}} 
\label{tab:ablation_module}
\end{table*}

\paragraph{Explored Paradigm} To evaluate the effects of various components in the proposed module, we conduct an ablation study on the main modules in Endo3DAC. In table~\ref{tab:ablation_module}, we present the results of the ablation study on four main components. \xmark \ in GDC-LoRA(Depth) and GDV-LoRA(Pose\&Intrinsic) represents using vanilla LoRA to replace our proposed module and \xmark \ in Con Neck and Self-SSI simply refers to disabling them.

\begin{figure}[t]
\centering
\includegraphics[width=0.95\linewidth]{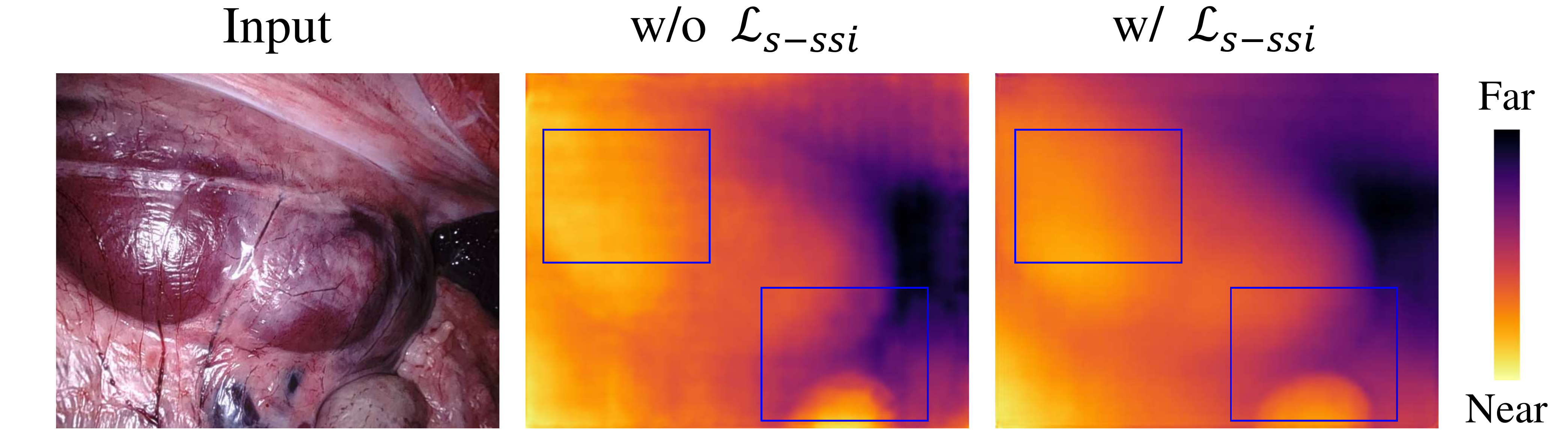}
\caption{An example to show the function of $\mathcal{L}_{s-ssi}$. Without $\mathcal{L}_{s-ssi}$, the small square segment problem occurs and the edges are blurry. When $\mathcal{L}_{s-ssi}$ is applied, surface depth values are smoother and the edges are clearer.}
\label{fig:sssivis}
\end{figure}

As expected, utilizing all the modules has the highest performance. By comparing the first four rows, each module has a positive effect on the performance, where applying GDV-LoRA for depth estimation increases the performance most, demonstrating the effectiveness of the fine-tuning strategy based on a pre-trained depth estimation foundation model. Adding GDV-LoRA also results in an improvement. This is because the estimation of relative poses is more accurate thus improving the reliability of reprojection constraint. Implementing Conv Neck also increases the RMSE by about 0.261 where the vision features in high-frequency areas are better captured. The proposed self-supervised scale- and shift-invariant loss improve the performance by a small margin, where its effects lie more on comprehensive and qualitative aspects. An example to show the function of $\mathcal{L}_{s-ssi}$ is presented in Fig.~\ref{fig:sssivis}. We can notice in the left blue square, we notice that without $\mathcal{L}_{s-ssi}$, the depth values are segmented with small squares, indicating the unevenness of depth values. The edges are blurred and not aligned with the input image in the right blue square without $\mathcal{L}_{s-ssi}$. $\mathcal{L}_{s-ssi}$ enforces the depth value to be consistent from different views leading to smoother surfaces and clearer edges.

\paragraph{Adaptation Method} We compare our proposed GDV-LoRA with three other adaptation methods including LoRA~\citep{hu2021lora}, AdaLoRA~\citep{zhang2023adaptive} and VeRA~\citep{kopiczko2024vera} to demonstrate the effectiveness of our fine-tuning method. As shown in Table~\ref{tab:ablation_lora}, our proposed GDV-LoRA results in the best performance compared to other methods. GDV-LoRA outperforms LoRA and AdaLoRA with less GPU Memory. While VERA utilizes the same memory usage as GDV-LoRA because of similar additional vector multiplication, we find that it results in significant decreases in all evaluation metrics. Fine-tuning only vectors while maintaining random metrics initialization is not capable of such dense estimation tasks. By contrast, our dynamic fine-tuning strategy enables the metrics to obtain informative initialization for subsequent vector-based fine-tuning.

\begin{table}[t]
\caption{Ablation study on different adaptation methods. }
\fontsize{10}{15}\selectfont
\centering
\resizebox{0.48\textwidth}{!}{
\begin{tabular}{c|c|ccccc}
\hline 
Methods & GPU Memory $\downarrow$ & Abs Rel $\downarrow$ & Sq Rel $\downarrow$ & RMSE $\downarrow$ & RMSE log $\downarrow$ & $\delta \uparrow$ \\ \hline 
VERA~\citep{kopiczko2024vera} & \textbf{19.82GB} &  0.072 & 0.458 & 5.169 & 0.090 & 0.965 \\  
LoRA~\citep{hu2021lora} & 22.38GB & 0.050 & 0.316 & \underline{4.175} & \underline{0.067} & \underline{0.982} \\  
AdaLoRA~\citep{zhang2023adaptive}  & \underline{21.76GB} & \underline{0.049} & \underline{0.315} & 4.181 & \underline{0.067} & \underline{0.982} \\  
GDV-LoRA (Ours) & \textbf{19.82GB} & \textbf{0.046} & \textbf{0.307} & \textbf{4.063} & \textbf{0.065} & \textbf{0.985} \\  \hline 

\end{tabular}} 
\label{tab:ablation_lora}
\end{table}

\paragraph{Foundation Models Size} We further carry out an ablation study on the size of pre-trained foundation models. We test by fine-tuning three weights provided by DA in their repository: Small (24.8M), Base (97.5M), and Large (335.3M). The results are shown in table~\ref{tab:ablation_size}. The performance increases but the inference speed decreases with the increase of size. The improvement conducted from Base to Large is very small with more than three times larger size and two times lower speed. Therefore, we choose Base for our proposed method for a compromise of performance, size, and speed.

\begin{table}[t]
\caption{Ablation study on the size of pre-trained foundation models. }
\fontsize{10}{15}\selectfont
\centering
\resizebox{0.48\textwidth}{!}{
\begin{tabular}{c|c|ccccc}
\hline 
Size & Speed. (ms) & Abs Rel $\downarrow$ & Sq Rel $\downarrow$ & RMSE $\downarrow$ & RMSE log $\downarrow$ & $\delta \uparrow$ \\ \hline 
Small (24.8M) & 6.1 &  0.050 & 0.334 & 4.376 & 0.066 & 0.984 \\  
Base (97.5M)  & 12.0 &  0.046 & 0.307 & 4.063 & 0.065 & 0.985 \\  
Large (335.3M)  & 22.5 &  0.045 & 0.302 & 4.012 & 0.064 & 0.987 \\  \hline 

\end{tabular}} 
\label{tab:ablation_size}
\end{table}

\subsection{Experiments on Pose and Intrinsic}

\subsubsection{Evaluation on SCARED and SimCol}

\begin{figure*}[t!]
\centering
\includegraphics[width=0.95\linewidth]{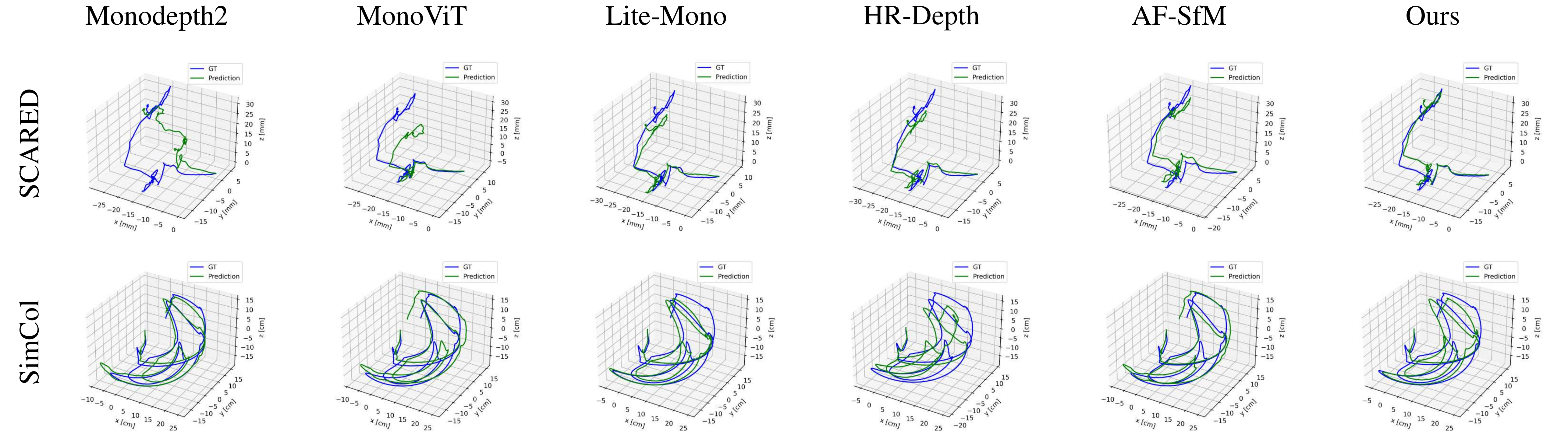}
\caption{Qualitative pose estimation comparison on \textbf{SCARED dataset} and \textbf{SimCol dataset}.}
\label{fig:posevis}
\end{figure*}

Following previous works~\citep{shao2022self, yang2024self, cui2024surgical}, we select two sequences in the SCARED dataset, and use all test sequences in the SimCol dataset to perform the 5-frame pose evaluation. Table~\ref{tab:pose_scared} and table~\ref{tab:pose_simcol} show the comparison of the proposed method with the other six methods. Our method achieves the lowest ATE and RPE in both datasets. Note that the improvement in pose estimation is not as much as in the depth estimation. This is because pose estimation can utilize the whole feature vector with all pixels to predict poses while depth estimation is a pixel-wise task, making pose estimation more robust to observation noise. Most of the previous utilize the same pose estimation network where a ResNet-18 network is used for the encoder with a separate convolutional decoder. A qualitative comparison is shown in Fig.~\ref{fig:posevis} where one sequence is selected from SCARED and SimCol, respectively. We can see that the trajectories of ours are better than those of the compared methods.

\begin{table}[t]
\caption{Quantitative pose estimation comparison on \textbf{SCARED dataset}. The ATE is averaged over all 5-frame snippets.}
\centering
\label{tab:pose_scared}  
\resizebox{0.48\textwidth}{!}{
\begin{tabular}{c|c|c|c}
\toprule  
Method & ATE $\downarrow$ (Seq.1) & ATE $\downarrow$ (Seq.2) & Mean ATE $\downarrow$\\ \midrule 
Monodepth2~\citep{godard2019digging} &  0.0798 & 0.0560 & 0.0662\\
MonoViT~\citep{zhao2022monovit} &  0.0765 & 0.0510 & 0.0638\\
HR-Depth~\citep{bian2019unsupervised} &  0.0770 & 0.0502 & 0.0636\\
Endo-SfM~\citep{ozyoruk2021endoslam} &  0.0759 & 0.0500 & 0.0629\\ 
AF-SfM~\citep{shao2022self} &  0.0742 & 0.0478 & 0.0610\\ 
Lite-Mono~\citep{zhang2023lite} &  \underline{0.0733} & \underline{0.0477} & \underline{0.0605} \\ \midrule 
\textbf{Endo3DAC (Ours)} &  \textbf{0.0724} & \textbf{0.0442} & \textbf{0.0583} \\ \bottomrule
\end{tabular}}
\end{table}

\begin{table}[t!]
\caption{Quantitative pose estimation comparison on \textbf{SimCol dataset}. The ATE and RPE are averaged over all 5-frame snippets.}
\centering
\label{tab:pose_simcol}  
\resizebox{0.48\textwidth}{!}{
\begin{tabular}{c|cc}
\toprule  
Method & ATE $\downarrow$ & RPE $\downarrow$ \\ \midrule 
MonoViT~\citep{zhao2022monovit} & 0.0156 $\pm$ 0.0126 & 0.0090 $\pm$ 0.0075 \\
Endo-SfM~\citep{ozyoruk2021endoslam} & 0.0154 $\pm$ 0.0121 & 0.0078 $\pm$ 0.0077 \\
AF-SfM~\citep{shao2022self} & 0.0150 $\pm$ 0.0113 & \underline{0.0081 $\pm$ 0.0060} \\
HR-Depth~\citep{lyu2021hr} & 0.0147 $\pm$ 0.0138 & 0.0094 $\pm$ 0.0081 \\
Monodepth2~\citep{godard2019digging} & 0.0146 $\pm$ 0.0129 & 0.0092 $\pm$ 0.0074 \\ 
Lite-Mono~\citep{zhang2023lite} & \underline{0.0144 $\pm$ 0.0114} & 0.0082 $\pm$ 0.0059 \\ \midrule 
\textbf{Endo3DAC (Ours)} & \textbf{0.0143 $\pm$ 0.0112} & \textbf{0.0079 $\pm$ 0.0053} \\  \bottomrule
\end{tabular}}
\end{table}

\subsubsection{Effectiveness of Integrated Network}

To validate the effectiveness of our proposed method where depth maps, poses, and camera intrinsics are estimated within an integrated network, we make a comparison experiment with traditional separate network methods. The results are shown in Table~\ref{tab:ablation_pose}. We use a separate ResNet-18-based Pose-Net as a comparison. Our proposed integrated framework has better performance in both ATE and RPE. The relative pose between two frames has strong correspondence with the depth variation therefore a pre-trained depth estimation foundation model should benefit pose estimation. By fine-tuning a small number of trainable parameters, the vision feature extracted by the network is more comprehensively utilized and optimized, resulting in higher accuracy in pose estimation.

\begin{table}[t!]
\caption{Comparison experiment between integrated network and separate Depth-Net and Pose-Net. We use a ResNet-18 for the separate Pose-Net comparison. }
\centering
\resizebox{0.40\textwidth}{!}{
\begin{tabular}{c|cc}
\toprule 
Pose-Net Architecture & ATE $\downarrow$ & RPE $\downarrow$ \\ \midrule 
Separate Pose-Net & 0.0151 $\pm$ 0.0119 & 0.0085 $\pm$ 0.0057 \\ 
Integrated (Ours) & 0.0143 $\pm$ 0.0112 & 0.0079 $\pm$ 0.0053 \\  \bottomrule

\end{tabular}} 
\label{tab:ablation_pose}
\end{table}

\subsubsection{Evaluation of Intrinsic Estimation}
To evaluate the accuracy of our proposed estimation method for camera intrinsic, we compare our method with two other methods~\citep{gordon2019depth, varma2022transformers} on the selected sequences of the SCARED dataset. As shown in table~\ref{tab:intrinsic}, our method has the lowest Abs Rel in all four camera intrinsic parameters. The accurate estimation of intrinsic will also benefit subsequent applications such as scene reconstruction.

\begin{table}[t!]
\caption{Quantitative intrinsic estimation comparison on \textbf{SCARED dataset}. The results are averaged on all selected frames. }
\fontsize{6}{8}\selectfont
\centering
\resizebox{0.40\textwidth}{!}{
\begin{tabular}{c|cccc}
\toprule 
\multirow{2}{*}{Method} & \multicolumn{4}{c}{Abs Rel $\downarrow$} \\ 
 & $f_x$ & $f_y$ & $c_x$ & $c_y$ \\ \midrule 
~\citep{gordon2019depth} & 0.028 & 0.030 & 0.027 & \underline{0.046} \\
~\citep{varma2022transformers} & \underline{0.020} & \underline{0.025} & \underline{0.023} & 0.054 \\
Ours & \textbf{0.001} & \textbf{0.014} & \textbf{0.002} & \textbf{0.029} \\  \bottomrule

\end{tabular}} 
\label{tab:intrinsic}
\end{table}

\subsection{Experiments on Scene Reconstruction}

\subsubsection{Evaluation on SCARED}

\begin{figure}[t]
\centering
\includegraphics[width=0.9\linewidth]{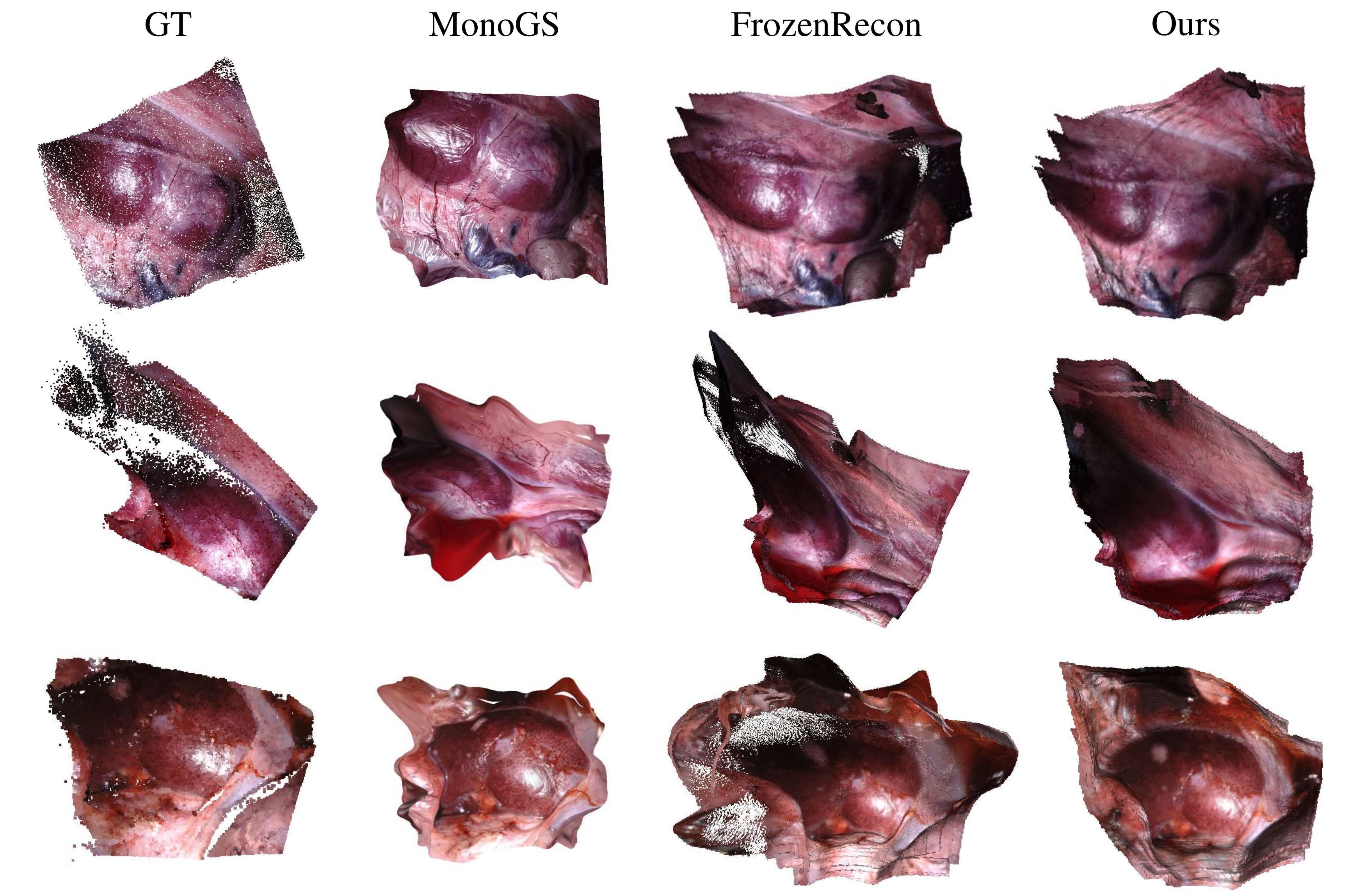}
\caption{Qualitative scene reconstruction comparison on \textbf{SCARED dataset}. Our method can generate better 3D scene shapes without any ground truth geometry or camera information.}
\label{fig:reconvis}
\end{figure}

To show the robustness and accuracy of our reconstruction method, we compared it with a SLAM-based method (MonoGS~\citep{Matsuki:Murai:etal:CVPR2024}) and a depth estimation-based reconstruction method (FrozenRecon~\citep{xu2023frozenrecon}). The quantitative results are shown in table~\ref{tab:recon_main}. Our method achieves the best results in most metrics except for Recall. In Fig.~\ref{fig:reconvis}, we show qualitative results compared to other methods. We observe that our method preserves more details while maintaining the completeness of the whole surface. The surfaces in our reconstructions are consistent and smooth without the hollows in FrozenRecon or the overly sharp edges in MonoGS.

\begin{table}[t!]
\caption{Qualitative comparison on the SCARED dataset of 3D scene reconstruction with state-of-the-art methods. We evaluate the Accuracy Acc, Completeness Comp, Chamfer distance Cham, and F-score F1 with a threshold of 5mm. }
\centering
\label{tab:recon_main}  
\resizebox{0.48\textwidth}{!}{
\begin{tabular}{c|cccccc}
\toprule
Method & Acc $\downarrow$ & Comp $\downarrow$ & Cham $\downarrow$ & Prec $\uparrow$ & Rec $\uparrow$ & F1 $\uparrow$ \\ \midrule 
MonoGS~\citep{Matsuki:Murai:etal:CVPR2024} &  \underline{3.56}  & 2.54  &  \underline{2.97}  &  70.0 &  81.4  &  75.2   \\
FrozenRecon~\citep{xu2023frozenrecon} & 4.91 & \underline{2.59} & 3.25 & \underline{74.0} & \textbf{90.7} &   \underline{77.5} \\ 
Ours & \textbf{3.08} & \textbf{1.86} & \textbf{2.66}  & \textbf{77.7}  & \underline{88.3}  & \textbf{82.6}  \\

\bottomrule

\end{tabular}}
\end{table}

\subsubsection{Ablation Study}

In table~\ref{tab:recon_abl}, we compared ablations of the main proposed components of our method. We can observe that decreases in metrics occur in any removal of patch-sampling strategy, appearance flow calibration, or pose and $K$ initialization. The performance abates the least without appearance flow calibration because of the robustness of the optimization process of the baseline method. Discarding the estimated pose and camera intrinsic $K$ degrades the performance the most. An accurate camera intrinsic is important for scene reconstruction because it influences the projection relationship of all frames which demonstrates the significance of our proposed network.

\begin{table}[t!]
\caption{Ablation study on the patch-sampling strategy, appearance flow calibration, and pose intrinsic initialization. }
\centering
\label{tab:recon_abl}  
\resizebox{0.48\textwidth}{!}{
\begin{tabular}{c|cccccc}
\toprule
Method & Acc $\downarrow$ & Comp $\downarrow$ & Cham $\downarrow$ & Prec $\uparrow$ & Rec $\uparrow$ & F1 $\uparrow$ \\ \midrule 
Ours w/o patch-sampling & 3.23 & 2.31 & 2.83 &  75.8  &  86.9   & 81.6 \\
Ours w/o appearance calib & 3.15 & 2.25 & 2.74 &  77.0  &  87.0   & 81.9 \\
Ours w/o pose \& $K$ init & 3.53 & 2.52 & 3.08 &  73.4  &  85.2   & 81.0 \\
Ours & 3.08 & 2.23 & 2.66 & 77.7 & 88.3 & 82.6 \\

\bottomrule

\end{tabular}}
\end{table}

\section{Conclusion}
In this article, we proposed a unified method for endoscopic scene reconstruction by efficiently adapting pre-trained foundation models. We first design a framework for endoscopic self-supervised depth estimation where depth map, relative pose, and camera intrinsic parameters are estimated with an integrated network. GDV-LoRA is designed to effectively fine-tune the model for different tasks with a small portion of parameters where most parameters are frozen during training. We further propose a pipeline for dense 3D scene reconstruction with our depth estimation network by jointly optimizing scale, shift of depth map, and dozens of camera parameters. Extensive experiments made on four publicly available datasets demonstrate the superiority and generalization ability of our method. Our method can be adapted to a variety of datasets because only endoscopic videos are required for training and evaluation. Our work also reveals the potentiality of leveraging existing foundation models to different domains with very few training parameters and computational resources.

\section{Acknowledgment}
This work was supported by Hong Kong Research Grants Council (RGC) Collaborative Research Fund (C4026-21G), General Research Fund (GRF 14211420 \& 14203323),  Shenzhen-Hong Kong-Macau Technology Research Programme (Type C) STIC Grant SGDX20210823103535014 (202108233000303), Regional Joint Fund Project of Guangdong Basic and Applied Research Fund 2021B1515120035 (B.02.21.00101).

\appendix
\section{Evaluation Metrics}
\label{app1}

For depth evaluation metrics, we report Abs Rel, Sq Rel, RMSE, RMSE log, and $\delta$. Their definitions can be found in Table~\ref{tab:app_em_depth}.

\begin{table}[!ht]
\caption{Definition of depth evaluation metrics. $D$ and $D^*$ are the predicted and ground truth depth maps.}
\fontsize{8}{10}\selectfont
\centering
\resizebox{0.40\textwidth}{!}{
\begin{tabular}{c|c}
\toprule 
Metrics Name & Definition \\ \midrule 
$AbsRel$ & $\frac{1}{\left| \textbf{D}\right|}\sum_{d\in \textbf{D}}^{}\left|d^{*}-d \right|/d^{*}$ \\ 
$SqRel$ & $\frac{1}{\left| \textbf{D}\right|}\sum_{d\in \textbf{D}}^{}\left|d^{*}-d \right|^{2}/d^{*}$ \\ 
$RMSE$ & $\sqrt{\frac{1}{\left| \textbf{D}\right|}\sum_{d\in \textbf{D}}\left| d^{*}- d\right|^{2}}$ \\
$RMSElog$ & $\sqrt{\frac{1}{\left| \textbf{D}\right|}\sum_{d\in \textbf{D}}\left| logd^{*}- logd\right|^{2}}$ \\ 
$\delta$ & $\frac{1}{\left| \textbf{D}\right|}\left| \left\{ d\in \textbf{D}|max(\frac{d^{*}}{d}, \frac{d}{d^{*}} < 1.25) \right\} \right| \times 100\%$  \\ \bottomrule 
\end{tabular}} 
\label{tab:app_em_depth}
\end{table}

For reconstruction evaluation metrics, we report Accuracy (Acc), Completeness (Comp), Chamfer distance (Cham), Precision (Prec), Recall (Rec), and F1-Score (F1). Their definitions can be found in Table~\ref{tab:app_em_recon}.

\begin{table}[!ht]
\caption{Definition of reconstruction evaluation metrics. $P$ and $P^*$ are the point clouds sampled from predicted and ground truth mesh.}
\fontsize{8}{10}\selectfont
\centering
\resizebox{0.40\textwidth}{!}{
\begin{tabular}{c|c}
\toprule 
Metrics Name & Definition \\ \midrule 
Acc & $\frac{1}{|P|}\left(\min _{p^* \in P^*}\left\|p-p^*\right\|\right)$ \\ 
Comp & $\frac{1}{|P^*|}\left(\min _{p \in P}\left\|p-p^*\right\|\right)$ \\ 
Cham &  $\frac{Acc + Comp}{2}$ \\
Prec & $\frac{1}{|P|}\left(\min _{p^* \in P^*}\left\|p-p^*\right\| < 0.05 \right)$ \\ 
Rec & $\frac{1}{|P^*|}\left(\min _{p \in P}\left\|p-p^*\right\| < 0.05 \right)$ \\ 
F1 & $\frac{2 \times Prec \times Rec}{Prec + Rec}$ \\ \bottomrule 
\end{tabular}} 
\label{tab:app_em_recon}
\end{table}

\section{More details on Network}
\begin{figure}[t]
\centering
\includegraphics[width=0.6\linewidth]{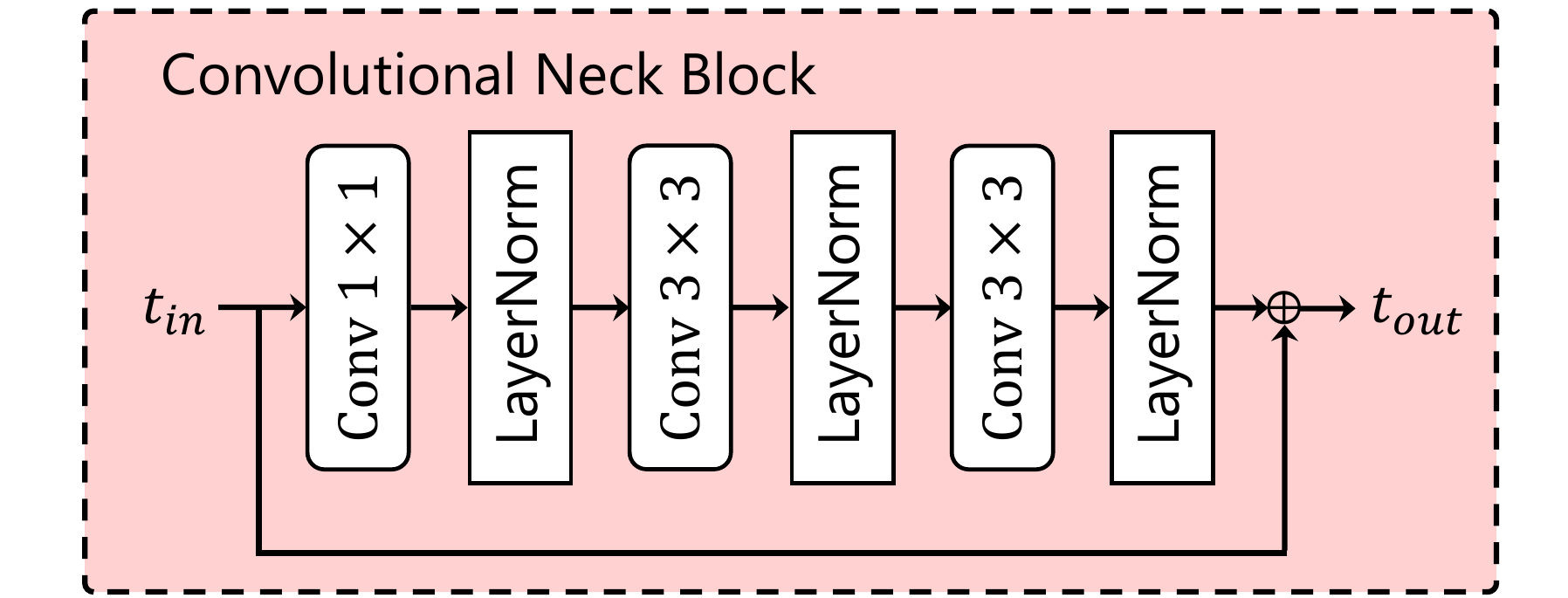}
\caption{Illustration of Convolution Neck block.}
\label{fig:block}
\end{figure}

The Convolution Neck block consists of three convolutional layers with LayerNorm and a residual connection to feed forward the results of transformer blocks, as shown in Figure~\ref{fig:block}. During training, we also incorporate an appearance flow network and an optical flow network proposed in~\citep{shao2022self} to address the issue of inconsistent lighting. Note that the appearance flow network and optical flow network only apply to calibrate warping images but are not considered part of the proposed network.

\section{More details on Baseline Settings}
All baseline methods are implemented based on their original released codes. Each output depth map is aligned with the ground truth depth map on both scale and shift before evaluation for a fair comparison. The maximum value set for each dataset was chosen either based on most previous baselines or the recommended setting given by the dataset's authors.

\bibliographystyle{elsarticle-harv} 
\bibliography{mybib}{}

\end{document}